\documentclass{article}

\usepackage[main, final]{neurips_2026}
\usepackage[utf8]{inputenc} % allow utf-8 input
\usepackage[T1]{fontenc}    % use 8-bit T1 fonts
\usepackage{hyperref}       % hyperlinks
\usepackage{url}            % simple URL typesetting
\usepackage{booktabs}       % professional-quality tables
\usepackage{amsfonts}       % blackboard math symbols
\usepackage{nicefrac}       % compact symbols for 1/2, etc.
\usepackage{microtype}      % microtypography
\usepackage{xcolor}         % colors

\usepackage{amsmath}
\usepackage[most]{tcolorbox}
\tcbuselibrary{breakable}
\definecolor{myblue}{RGB}{0, 102, 204}
\definecolor{lightblue}{RGB}{210, 225, 245}
\usepackage{cleveref}

\usepackage{wrapfig}      
\usepackage{booktabs}
\usepackage{tabularx}
\usepackage{xcolor}
\usepackage{threeparttable}
\usepackage{adjustbox}

\usepackage{multirow}   % 提供 \multirow
\usepackage[table]{xcolor}

\usepackage{rotating}
\usepackage{float}
\usepackage{wrapfig}

\title{Learning from the Self-future: \\
On-policy Self-distillation for dLLMs}

\author{%
  Yifu Luo $^{1, \dagger}$,
  Zeyu Chen $^{2, \dagger}$,
  Haoyu Wang $^{3}$,
  Xinhao Hu $^{1}$,
  \AND
  \textbf{Yuxuan Zhang} $^{4}$,
  \textbf{Zhizhou Sha} $^{5}$,
  \textbf{Shiwei Liu} $^{6, 7, 8, }$ \thanks{Corresponding Author.}
  \\
  $^{\dagger}$Equal Contribution 
  \vspace{4pt}
  \\
  $^{1}$Tsinghua University \vspace{.1em} \hspace{4pt} $^{2}$Technical University of Munich \vspace{.1em} \hspace{4pt} $^{3}$Nanyang Technological University \vspace{.1em} \hspace{4pt}
  \\
  $^{4}$University of British Columbia \vspace{.1em} \hspace{4pt} $^{5}$University of Texas at Austin \vspace{.1em} \hspace{4pt} 
  $^{6}$ELLIS Institute Tübingen \vspace{.1em} \hspace{4pt}
  \\
  $^{7}$Max Planck Institute for Intelligent Systems \vspace{.1em} \hspace{4pt}
  $^{8}$Tübingen AI Center \vspace{.1em} \hspace{4pt}
}

\begin{document}

\newcommand{\question}[2]{%
\begin{tcolorbox}[
    enhanced,
    colback=white,
    colframe=white,
    leftrule=0.4mm,
    rightrule=0.4mm,
    toprule=0.4mm,
    bottomrule=0.4mm,
    arc=0mm,
    left=0pt,
    right=0pt,
    top=2pt,
    bottom=2pt,
    breakable,
    borderline north={0.4mm}{0pt}{myblue},
    borderline south={0.4mm}{0pt}{myblue}
]
{\textbf{\textcolor{myblue}{Question:}}} #1 \\
{\textbf{\textcolor{myblue}{Answer:}}} #2
\end{tcolorbox}
}

\newcommand{\twoquestion}[4]{%
\begin{tcolorbox}[
    enhanced,
    colback=white,
    colframe=white,
    leftrule=0.4mm,
    rightrule=0.4mm,
    toprule=0.4mm,
    bottomrule=0.4mm,
    arc=0mm,
    left=0pt,
    right=0pt,
    top=2pt,
    bottom=2pt,
    breakable,
    borderline north={0.4mm}{0pt}{myblue},
    borderline south={0.4mm}{0pt}{myblue}
]
{\textbf{\textcolor{myblue}{Question:}}} #1 \\
{\textbf{\textcolor{myblue}{Answer:}}} #2 \\
\\
{\textbf{\textcolor{myblue}{Question:}}} #3 \\
{\textbf{\textcolor{myblue}{Answer:}}} #4
\end{tcolorbox}
}

\newcommand{\multiquestion}[8]{%
\begin{tcolorbox}[
    enhanced,
    colback=white,
    colframe=white,
    leftrule=0.4mm,
    rightrule=0.4mm,
    toprule=0.4mm,
    bottomrule=0.4mm,
    arc=0mm,
    left=0pt,
    right=0pt,
    top=2pt,
    bottom=2pt,
    breakable,
    borderline north={0.4mm}{0pt}{myblue},
    borderline south={0.4mm}{0pt}{myblue}
]
{\textbf{\textcolor{myblue}{Question:}}} #1 \\
{\textbf{\textcolor{myblue}{Answer:}}} #2 \\
\\
{\textbf{\textcolor{myblue}{Question:}}} #3 \\
{\textbf{\textcolor{myblue}{Answer:}}} #4 \\
\\
{\textbf{\textcolor{myblue}{Question:}}} #5 \\
{\textbf{\textcolor{myblue}{Answer:}}} #6 \\
\\
{\textbf{\textcolor{myblue}{Question:}}} #7 \\
{\textbf{\textcolor{myblue}{Answer:}}} #8
\end{tcolorbox}
}

\newcommand{\mask}{{\texttt{mask}}}

\maketitle

\begin{abstract}
On-policy self-distillation (OPSD) has proven effective for post-training large language models (LLMs), yet its application to diffusion LLMs (dLLMs) remains unexplored. Existing OPSD methods are inherently autoregressive-centric. They inject privileged information via left-to-right prefix conditioning with token-level divergence supervision, a design that fundamentally conflicts with the arbitrary-order generation of dLLMs. We introduce d-OPSD, the first OPSD framework tailored for dLLMs. Our approach makes two core contributions. First, we reframe self-teacher construction by using self-generated answers as suffix conditioning, enabling the student model to learn from ``self future-experience'' rather than privileged prefixes. Second, we shift supervision from token-level to step-level, aligning training with the iterative denoising process of dLLMs. Experiments across four reasoning benchmarks show that d-OPSD consistently outperforms RLVR and SFT baselines with superior sample efficiency, requiring only around $10\%$ of the optimization steps by RLVR and opening a promising pathway for dLLM post-training. The code is available at \url{https://github.com/xingzhejun/d-OPSD}.
\end{abstract}

\section{Introduction}

\begin{wrapfigure}{r}{0.6\textwidth}
\vspace{-52pt}
\centering
\includegraphics[width=0.6\textwidth]{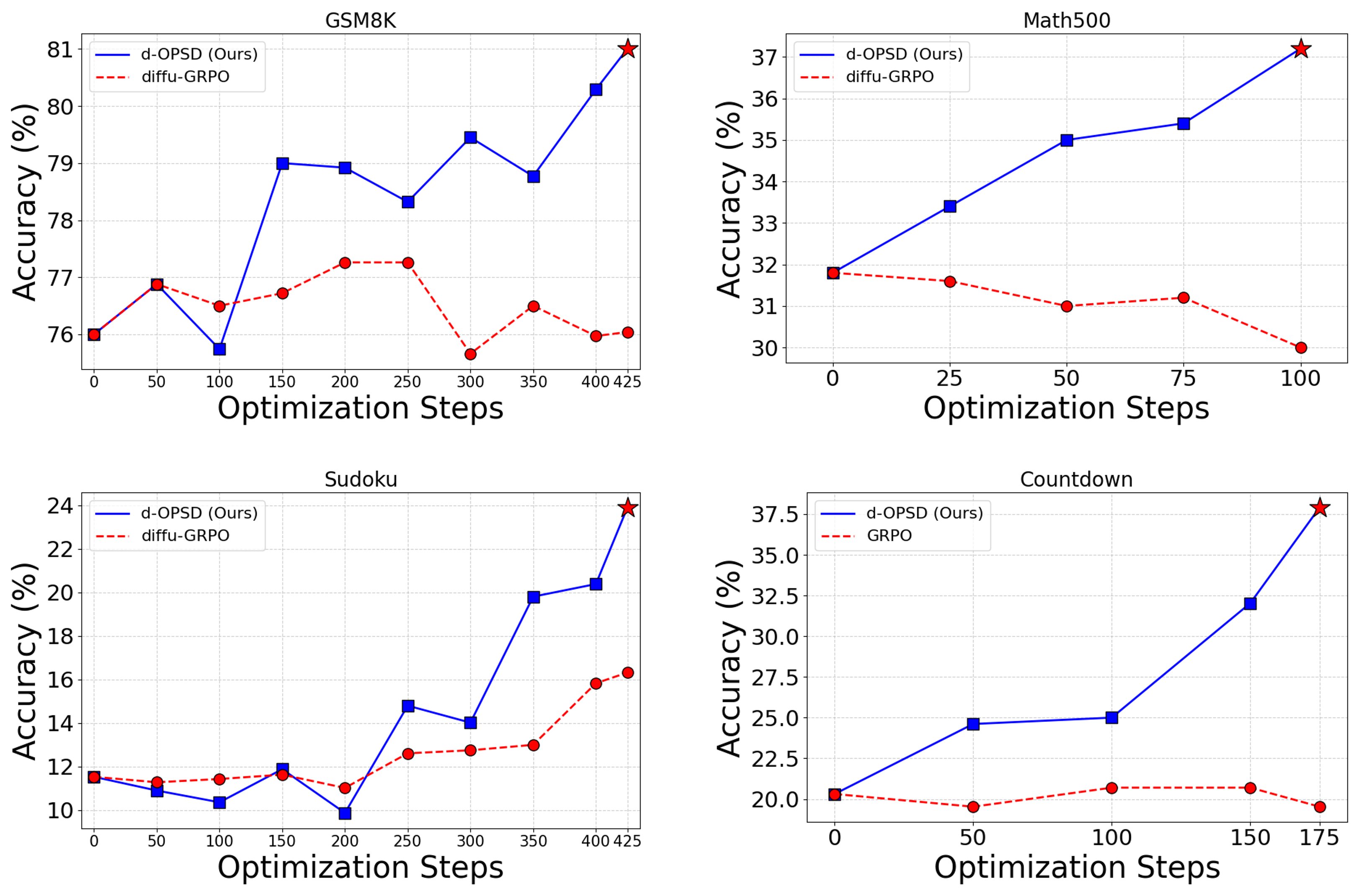}
\vspace{-14pt}
\caption{The reasoning performance and sample efficiency comparisons between the RLVR baseline (diffu-GRPO\citep{zhao2025d1}) and our approach, d-OPSD.}
\label{fig1}
\vspace{-20pt}
\end{wrapfigure}

On-policy distillation (OPD) \citep{agarwal2024policy,yang2025qwen3,lu2025onpolicydistillation,li2026rethinking}, where a student model samples its own trajectories while a stronger teacher model provides dense token-level supervision, has recently emerged as a highly effective paradigm for Large Language models (LLMs) post-training, offering significant advantages over Reinforcement Learning with Verifiable Rewards (RLVR) (e.g., GRPO \citep{guo2025deepseek} and supervised fine-tuning (SFT). Compared to RLVR, OPD provides dense token-level supervision from a teacher, overcoming the bottleneck of sparse outcome rewards. Compared to SFT, OPD utilizes generations sampled from the student itself, thereby preventing exposure bias \citep{bengio2015scheduled}. However, OPD relies heavily on a stronger teacher model, which is often impractical in many settings. To address this, recent works \citep{zhao2026self, hubotter2026reinforcement, shenfeld2026self} have extended OPD to on-policy self-distillation (OPSD), where a single model serves as its own teacher given teacher-specific privileged information, demonstrating a powerful framework for self-improvement.

Concurrently, diffusion large language models (dLLMs) \citep{ou2024your, nie2025large, ye2025dream, cheng2025sdar, bie2025llada2} have demonstrated strong potential as an alternative to autoregressive (AR) LLMs \citep{jaech2024openai, xiao2026mimo}. By modeling language generation as an iterative denoising process, dLLMs bypass the strict left-to-right dependency of AR models, unlocking unique advantages such as arbitrary-order generation and speed-up inference \citep{khanna2025mercury, song2025seed, wu2025fast}.

\begin{figure}
\centering{\includegraphics[scale=0.27]{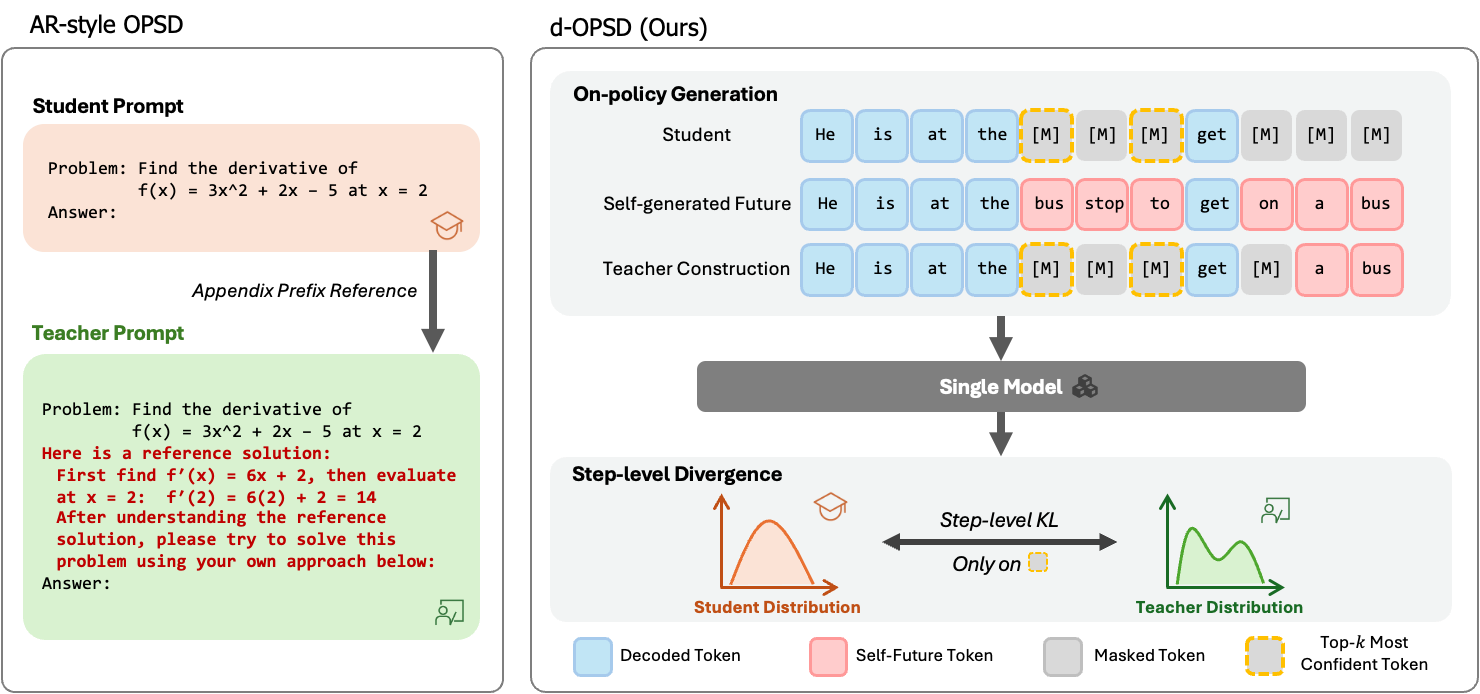}}
\vspace{-12pt}
\caption{The framework of our approach, d-OPSD. It leverages self-generated answers as suffix privileged information to construct the self-teacher, and uses step-level divergence to guide the student learn from the ``self future-experience''.}
\label{fig2}
\vspace{-12pt}
\end{figure}

While recent works \citep{zhao2025d1, tang2025wd1, xie2025step} has successfully applied RLVR to dLLMs demonstrating that their reasoning ability can be enhanced by post-training, OPSD for dLLMs remains largely unexplored in this context. Meanwhile, as shown in \Cref{fig2}, existing OPSD approaches for AR models follow a standard paradigm for self-teacher construction, where privileged information (e.g., reference solutions) is simply appended to the prompt, and teacher-student divergence supervision is calculated at the token level. Given that dLLMs exhibit fundamental different features from AR LLMs, we investigate the following two questions in this paper:

\twoquestion{Is there a better OPSD formulation tailored specifically for dLLMs?}{Yes. Both the self-teacher construction and the level of divergence supervision can be optimized for dLLMs, as shown in \Cref{fig2}.}
{Does OPSD outperform RLVR in enhancing the reasoning ability of dLLMs?}{Yes. It achieves superior results in both reasoning performance and sample efficiency, as shown in \Cref{fig1}.}

First, we identify that the self-teacher construction mentioned above is suboptimal for dLLMs. Appending privileged information to the prompt is inherently designed for AR models, because they are constrained to left-to-right generation where only prefix conditioning $p(\text{suffix} | \text{prefix})$ is available. In contrast, dLLMs generate sequences non-autoregressively, which allows us to incorporate privileged information as a suffix context condition. More importantly, this feature enables us to shift the content of privileged information from static reference solutions to the model's self-generated answers, adhering closer to the on-policy nature. As shown in \Cref{fig2}, the $p(\text{prefix} | \text{suffix})$ capability of dLLMs allows us to use self-generated answers as a suffix conditional posterior for privileged information. This guides the student to learn from ``self future-experience'', which is similar to human inspiration that we always daydream if we could go back to $10$ years ago knowing what happened next. A key advantage of our teacher construction is that it provides more new knowledge (thinking patterns) to transfer to the student, a claim we empirically discuss in \Cref{sec4.3}.

Second, token-level divergence supervision is not suitable for dLLMs either. While AR models natively rely on next-token prediction, dLLMs predict all masked tokens simultaneously at each denoising step, but only keep part of them while remasking others. Consequently, token-level supervision designed for AR models becomes incompatible. Instead, as each denoising step can be viewed as an independent markov transition, step-level divergence serves as a nature choice for dLLMs OPSD. By shifting the dense supervision from the token-level to the step-level, we closely align the OPSD objective with the iterative denoising nature of dLLMs.

Building on these insights, we propose \textbf{diffusion On-Policy Self-distillation (d-OPSD)}, a novel OPSD framework specifically designed for dLLMs to drive self-improvement. To the best of our knowledge, this represents the first application of OPSD to dLLMs. In our approach, the student samples its own trajectories, while the self-teacher is constructed using self-generated answers as suffix privileged information. By applying step-level divergence, the student effectively learns from its ``self future-experience''. Extensive experiments across four reasoning tasks demonstrate that our approach consistently outperforms RLVR and SFT baselines with superior reasoning performance and sample efficiency, as highlighted in \Cref{fig1}.

Our contributions are summarized as follows:

\begin{itemize}
    \item We identify that existing OPSD formulations are suboptimal for dLLMs. To bridge this gap, we introduce a novel self-teacher construction that utilizes self-generated answers as a suffix conditional posterior for privileged information, and we shift the dense divergence supervision from the token-level to the step-level.
    \item We are the first to introduce OPSD to dLLMs. We propose d-OPSD, a novel OPSD framework tailored for dLLMs to drive self-improvement. It enables a single model to act as both teacher and student, leveraging self-generated ``future'' as privileged information to provide dense step-level supervision over the student trajectories.
    \item We conduct extensive experiments across four reasoning tasks, demonstrating that our approach achieves both superior reasoning performance and sample efficiency compared to RLVR and SFT baselines. Furthermore, we empirically analyze the impact of different settings, paving the way for future advances in this field.  
\end{itemize}

\section{Preliminaries}

\subsection{Diffusion Large Language Models}

In this subsection, we briefly review the training and inference paradigms of dLLMs. During training, dLLMs define a forward process that gradually corrupts a clean input by replacing its tokens with a special $\mask$ token. Given a prompt $x$ and a clean response $y_0 = \{y_0^1, y_0^2, \cdots, y_0^L\}$, the forward process at step $0<t \leq T$ can be expressed as:
\begin{equation}
    q(y_t|y_0,x) = \prod_{i=1}^L q(y_t^i|y_0^i, x) \quad \text{and} \quad q_t(y_t^i|y_0^i, x) = 
\begin{cases} 
\frac{T - t}{T}, & y_t^i = y_0^i, \\
\frac{t}{T}, & y_t^i = \mask,
\end{cases}
\end{equation}
where L is the sequence length, and the superscript $i$ refers to the token position. 

In this work, we primarily focus on the reverse inference process of dLLMs. Given a prompt $x$ and a trained model $p_\theta$, inference is formulated as a $T$-step iterative denoising process, from a fully masked sequence $y_T = \{\mask\}^L$ to a clean response $y_0$. At each denoising step $t$, the model first computes the distribution for all tokens:
\begin{equation}
    \mathcal{P}_{t}^i = p_\theta(y^i|y_t, x), \quad 1 \leq i \leq L.
\end{equation}
For the top-$k$ most confident predictions among the currently masked positions, they are sampled and revealed. The remaining masked positions are kept masked as $\mask$ and to form $y_{t-1}$. After $T$ steps, all masked tokens are revealed, yielding the final response $y_0$. Additional preliminaries about block-diffusion, a common-used inference strategy, are provided in \Cref{app1.1}.

\subsection{On-policy Distillation}

OPD transfers knowledge from a stronger teacher model $p_T$ to a weaker student model $p_\theta$ by enforcing dense supervision over trajectories sampled by the student. For AR models, given a prompt $x$, the student samples a response $y = \{y^1, y^2, \cdots, y^L\}$. Using the AR factorization, the learning objective is to minimize the token-level KL between the teacher's and the student's next-token distributions:
\begin{equation}
    \mathcal{L}_\text{OPD} (\theta) = \mathbb{E}_x \left[ \sum_{i=1}^L \mathcal{D}_\text{KL} \left(p_\theta \left( \cdot | y^{<i}, x  \right) || \left( p_T \left( \cdot | y^{<i}, x \right) \right) \right) \right],
\label{eq4}
\end{equation}
where $p(\cdot | y^{<i}, x)$ denotes the distribution over the next token $y^i$. While we use reverse KL, forward KL and other distribution divergence measures like generalized Jensen-Shannon divergence \citep{agarwal2024policy} can also be employed.

Recent advances have extended OPD to OPSD, where the student and teacher are instantiated from the same model, denoted as $p_\theta$. The difference lies entirely in their conditioning contexts. For AR models, privileged information $y^*$, such as reference solutions \citep{zhao2026self} or environment feedback \citep{hubotter2026reinforcement}, is appended to the original prompt $x$ to construct a teacher-specific prompt $x^* = x + y^*$. Thus, the teacher distribution is:
\begin{equation}
    p_T = p_\theta \left( \cdot | y^{<i}, x, y^* \right) = p_\theta \left( \cdot | y^{<i}, x^* \right).
\end{equation}
Consequently, the learning objective in \Cref{eq4} adapts into the following:
\begin{equation}
    \mathcal{L}_\text{OPSD} (\theta) = \mathbb{E}_x \left[ \sum_{i=1}^L \mathcal{D}_\text{KL} \left(p_\theta \left( \cdot | y^{<i}, x  \right) || \left( p_\theta \left( \cdot | y^{<i}, x^* \right) \right) \right) \right].
\end{equation}

In this setup, both the teacher and student share the same model but differ only in the conditioning contexts, and the response is solely generated from the student. While OPSD achieves comparable performance to RLVR with superior sample efficiency for AR models, adapting this formulation to dLLMs presents fundamental challenges. First, the arbitrary-order generation of dLLMs provides an alterative for injecting privileged information, which better aligns with on-policy nature (\Cref{sec3.1}). Second, token-level divergence supervision is incompatible with dLLMs as next-token prediction is not factorized. Instead, step-level divergence supervision must be adopted(\Cref{sec3.2}).

\section{Methods}

\subsection{Teacher Construction: Learning from the Self-future} \label{sec3.1}

In this section, we describe how we utilize the student's self-generated ``future'' answers as privileged information for the teacher, which adheres closer to dLLMs and on-policy nature. While AR models are constrained to left-to-right generation with only $p(\text{suffix} | \text{prefix})$ available, dLLMs possess the bidirectional capability to model suffix conditioning $p(\text{prefix} | \text{suffix})$. As shown in \Cref{fig2}, our core insight is that after sampling a complete trajectory from the student, we can partially reveal this self-generated subsequent trajectory to the teacher as privileged information.

Specifically, We instantiate both the teacher and student distributions from the same dLLM $p_\theta$ by varying the conditioning inputs. Given a prompt $x$, the student first samples a trajectory from $p_\theta$:
\begin{equation}
    Y = \{ y_T, y_{T-1}, \cdots, y_0 \} \sim p_\theta (\cdot | x),
\label{eq7}
\end{equation}
where $y_T = \{ \mask \}^L$ is a fully masked sequence, $y_0$ is the final response, $T$ refers to the total number of denoising steps, and $L$ denotes the sequence length. At each denoising step $0<t \leq T$, the student input is simply the current noisy sequence:
\begin{equation}
    y_{\text{student},t} = y_t.
\label{eq8}
\end{equation}
Conversely, the teacher input is constructed by selectively revealing tokens from the final generated response $y_0$:
\begin{equation}
y_{\text{teacher},t}^i =
\begin{cases}
y_0^i, & \text{if } i \in \mathcal{S}_t,\\[4pt]
y_t^i, & \text{otherwise},
\end{cases}
\label{eq9}
\end{equation}
where $\mathcal{S}_t \subset \{ 1, 2, \cdots, L \}$ is the revealing subset of indices randomly selected with a fixed retaining ratio $\rho_\text{teacher}$ from the currently masked positions. Thus, both the student and teacher share the same model $p(\theta)$, but the teacher benefits from the self-generated ``future'' trajectory. An illustration example of our teacher construction is provided in \Cref{app2}. 

This construction seamlessly aligns with on-policy and dLLMs nature. First, all data is generated by the student. Second, the construction in \Cref{eq8} and \Cref{eq9} yields distributions $p_\theta(\cdot | y_{\text{student}, t}, x)$ and $p_\theta(\cdot | y_{\text{teacher}, t}, x)$, which enable a direct step-level divergence supervision, which we introduce in the next subsection. Note that $p(\cdot | y_{t}, x)$ here denotes distribution for the next step.

\subsection{Step-level Divergence Supervision} \label{sec3.2}

Unlike AR models, which natively employ token-level supervision via next-token prediction, dLLMs decode sequences via next-step prediction. At each denoising step, only the top-$k$ most confident tokens among the currently masked positions are sampled and revealed, while the remaining $\mask$ tokens are kept masked. While token-level supervision is incompatible, we propose step-level divergence supervision as a more natural objective for dLLMs.

Specifically, at each denoising step $t$, using the previously constructed inputs $y_{\text{student},t}$ and $y_{\text{teacher},t}$, the model first computes full-sequence distributions:
\begin{equation}
\begin{aligned}
    \mathcal{P}_{\text{student}, t}^i &= p_\theta(y^i|y_{\text{student},t}, x), \quad 1 \leq i \leq L, \\
    \mathcal{P}_{\text{teacher}, t}^i &= p_\theta(y^i|y_{\text{teacher},t}, x), \quad 1 \leq i \leq L.
\end{aligned}
\label{eq10}
\end{equation}
Crucially, not all token positions $i$ actively participate in the state transition from $t$ to $t-1$. We focus exclusively on the top-$k$ most confident tokens among the currently masked positions, as only these tokens dictate the step-level transition. Denoting these tokens' indices as the top-$k$ subset $\mathcal{K}_t \subset \{ 1 \leq i \leq L | y_t^i = \mask  \}$ which satisfies:
\begin{equation}
    \sum_{t=1}^T |\mathcal{K}_t|= L.
\end{equation}
We then compute the step-level KL divergence over this subset:
\begin{equation}
    \mathcal{L}_t = \frac{1}{|\mathcal{K}_t|}  \sum_{i \in \mathcal{K}_t}  \mathcal{D}_\text{KL} \left(  \mathcal{P}_{\text{student}, t}^i || \mathcal{P}_{\text{teacher}, t}^i \right).
\end{equation}
Note that the top-$k$ subset $\mathcal{K}_t$ can theoretically be determined from either the student distribution or teacher distribution. However, the ablation study in \Cref{sec4.4.3} suggests that deriving from the teacher distribution yields greater performance gains.

With the self-teacher construction and step-level divergence in place, we now possess all the essential components needed to apply OPSD to dLLMs.

\subsection{d-OPSD: the First On-Policy Self-distillation for dLLMs} \label{sec3.3}

We now formally introduce our approach, \textbf{d-OPSD}. Operating with a single model $p_\theta$ severing simultaneously as student and teacher, the procedure begins with the student sampling an on-policy $T$-step trajectory $Y$ (\Cref{eq7}) for a given prompt $x$. For each denoising step $0 <t \leq T$, we construct the student input $y_{\text{student},t}$ and the teacher input $y_{\text{teacher},t}$ using \Cref{eq8} and \Cref{eq9}. Note that the constructions are independent over steps. Finally, we minimize the following step-level learning objective across the entire on-policy trajectory:
\begin{equation}
\begin{aligned}
    \mathcal{L}_\text{OPSD} (\theta) =& \mathbb{E}_x \left[ \frac{1}{T} \sum_{t=1}^T \mathcal{L}_t \right] \\
    =& \mathbb{E}_x \left[ \frac{1}{T} \sum_{t=1}^T \frac{1}{|\mathcal{K}_t|}  \sum_{i \in \mathcal{K}_t}  \mathcal{D}_\text{KL} \left(  \mathcal{P}_{\text{student}, t}^i || \mathcal{P}_{\text{teacher}, t}^i \right) \right] \\
    =& \mathbb{E}_x \left[ \frac{1}{T} \sum_{t=1}^T \frac{1}{|\mathcal{K}_t|}  \sum_{i \in \mathcal{K}_t}  \mathcal{D}_\text{KL} \left(  p_\theta \left(y^i|y_{\text{student},t}, x \right) || p_\theta \left(y^i|y_{\text{teacher},t}, x \right) \right) \right].
\end{aligned}
\label{eq12}
\end{equation}

Additionally, we find that the quality of the student trajectory $Y$ influences the final performance (see \Cref{sec4.4.4} and \Cref{app5.2}). Therefore, for each prompt $x$, we keep sampling trajectories until a correct final answer $y_0$ occurs, or the sampling iteration number meets a threshold (similar to pass@$k$, and we set $k=8$ by default)\footnote{Even with $k=1$, our approach still surpasses the RLVR baseline which uses group $k=8$ rollouts, see \Cref{sec4.4.4}.}. Note that this sampling strategy shares the same computation overhead as RLVR (group $k$ rollouts) for each training step. Following \citep{zhao2026self}, we apply pointwise KL clipping and the fix teacher strategy, as detailed in \Cref{app3}. Additional implementation details are also provided in \Cref{app3}, including an important engineering technique preventing out of memory by concatenating step-level inputs, motivated by \citep{wang2025revolutionizing}.

Crucially, we conclude this section by highlighting the fundamental distinctions between our approach and existing self-distillation approaches for dLLMs, such as d3llm \citep{qian2026d3llm} and Cd4lm \citep{liang2026cd4lm}, which also construct a self-teacher by partially revealing answers. First and foremost, the revealed answers in our approach are ``self-experience'' generated on-policy by the student itself, whereas theirs are from the ground-truth of static datasets. Second, while we leverage step-level divergence supervision across an entire on-policy generation trajectory, they employ a single forward pass like a `one-step'' fake trajectory to provide supervision. These critical differences define d-OPSD as an on-policy distillation approach providing dense supervision for every denoising steps across the entire trajectory, whereas their approaches remain fundamentally off-policy closely related to SFT.

\section{Experiments} \label{sec4}

In this section, we first address a foundational prerequisite with a toy verification:

\question{Is the self-teacher strong enough to guide self-distillation?}{Yes. The self-teacher is capable enough that the correct answer can be resumed using our teacher construction. See \Cref{sec4.1}.}

We then conduct comprehensive experiments to answer the following core questions:

\multiquestion{How does OPSD compare to SFT and RLVR in reasoning performance and sample efficiency?}{It outperforms or matches SFT and RLVR baselines in reasoning performance, while demonstrating vastly superior sample efficiency. See \Cref{sec4.2}.}
{How does the self-teacher construction in d-OPSD compare to the AR-style counterpart (\Cref{fig2})?}{Our approach significantly outperforms the AR counterpart. The key reason is that our teacher construction introduces more new knowledge (thinking patterns) to transfer to the student. See \Cref{sec4.3}.}
{What is the impact of different training settings?}{We provide comprehensive ablation results on different KL objectives, retaining ratios $\rho_\text{teacher}$, top-$k$ subset $\mathcal{K}_t$ selections, sampling strategies, and other training settings. See \Cref{sec4.4} and \Cref{app5.1}.}
{What are the failure modes for d-OPSD?}{Similar to RLVR,  OPSD is susceptible to policy collapse after reaching peak performance. See \Cref{sec4.5}.}

\subsection{Experimental Setup \& Toy Verification} \label{sec4.1}

\textbf{Models and Tasks.} We employ LLaDA-8B-Instruct \citep{nie2025large}, a state-of-the-art dLLM that has not undergone post-training, as our base model \footnote{We did not use Dream \citep{ye2025dream} because its output format is highly inconsistent, which causes severe
instability across RLVR baselines. This limitation is also marked by \citep{pan2025d}.}. We conduct experiments across four reasoning tasks spanning two categories: mathematical reasoning and planning. The mathematical reasoning tasks include GSM8K \citep{cobbe2021training} and MATH500 \citep{lightman2023let}. The planning tasks include $4x4$ Sudoku puzzles, which require
constraint satisfaction to fill a grid with numbers, and Countdown ($3$ numbers), where models must reach a target number using basic arithmetic operations on a given set of integers. All datasets configurations remain consistent with the RLVR baseline, diffu-GRPO \citep{zhao2025d1}. 

\begin{table}[!t]
\centering
\caption{Reasoning performance comparison across four reasoning tasks. Results of diffu-GRPO and the SFT varient are sourced from the original paper \citep{zhao2025d1}. Results of VRPO, d3LLM and the base model are evaluated using their open-sourced models. d-OPSD consistently outperforms or matches SFT and RLVR baselines.}
\begin{threeparttable}
\begin{adjustbox}{max width=\textwidth}
\begin{tabular}{lcccccccc}
\toprule
\multirow{2}{*}{Model / Seq Len} & \multicolumn{2}{c}{GSM8K} & \multicolumn{2}{c}{MATH500} & \multicolumn{2}{c}{Countdown} & \multicolumn{2}{c}{Sudoku} \\
\cmidrule(lr){2-3} \cmidrule(lr){4-5} \cmidrule(lr){6-7} \cmidrule(lr){8-9}
 & 256 & 512 & 256 & 512 & 128 & 256 & 128 & 256 \\
\midrule

LLaDA-8B-Instruct & 76.0 & 79.5 & 31.8 & 36.2 & 20.3 & 19.1 & 11.5 & 6.9 \\
\midrule
& \multicolumn{8}{c}{\textbf{SFT}} \\
SFT Varient \citep{zhao2025d1} & 78.8 & 81.1 & 32.6 & 34.8 & 20.3 & 14.5 & 16.5 & 8.5 \\
d3LLM \citep{qian2026d3llm} & 72.7 & 76.4 & 30.6 & 37.0 & 36.7 & 12.5 & 9.1 & 6.6 \\
\midrule
& \multicolumn{8}{c}{\textbf{RLVR}} \\
diffu-GRPO \citep{zhao2025d1} & 79.8 & 81.9 & 37.2 & \cellcolor{lightblue} 39.2 & 33.2 & 31.3 & 18.4 & 12.9 \\
VRPO \citep{zhu2025llada} & 80.1 & 81.5 & 35.6 & 34.8 & 21.9 & 21.1 & 12.8 & 9.6 \\
\midrule
& \multicolumn{8}{c}{\textbf{OPSD}} \\
\textbf{d-OPSD (Ours)} & \cellcolor{lightblue} 81.0 & \cellcolor{lightblue} 82.2 & \cellcolor{lightblue} 37.2 & 37.8 & \cellcolor{lightblue} 37.9 & \cellcolor{lightblue} 32.3 & \cellcolor{lightblue} 23.9 & \cellcolor{lightblue} 20.6 \\

\bottomrule
\label{tab2}
\end{tabular}
\end{adjustbox}
\end{threeparttable}
\vspace{-20pt}
\end{table}

\begin{wraptable}{r}{0.65\textwidth}
\vspace{-18pt}
\setlength{\tabcolsep}{4pt}
\centering
\caption{Sample efficiency comparison between the RLVR baseline and our approach. The optimization steps for diffu-GRPO are sourced from the original paper \citep{zhao2025d1}.}
\begin{tabular}{lcccc}
\toprule
\textbf{Method / Step} & \textbf{GSM8K} & \textbf{Math500} & \textbf{Countdown} & \textbf{Sudoku} \\
\midrule
diffu-GRPO & 7700 & 6600 & 5000 & 3800 \\
\textbf{d-OPSD (Ours)} & \cellcolor{lightblue} 425 & \cellcolor{lightblue}  100 & \cellcolor{lightblue}  175 & \cellcolor{lightblue}  425\\
\bottomrule
\label{tab3}
\end{tabular}
\vspace{-15pt}
\end{wraptable}

\textbf{Baselines. } We compare against two categories of post-training methods: RLVR and SFT. RLVR baselines include diffu-GRPO \citep{zhao2025d1} and VRPO \citep{zhu2025llada}. For SFT, we compare against the SFT variant from \citep{zhao2025d1} and the existing off-policy self-distillation approach, d3LLM \citep{qian2026d3llm}.

\begin{wraptable}{r}{0.65\textwidth}
\vspace{-15pt}
\setlength{\tabcolsep}{4pt}
\centering
\caption{Toy Verification. The correct answer can be resumed from the self-teacher construction.}
\small
\begin{tabular}{lcccc}
\toprule
\textbf{Method / Tasks} & \textbf{GSM8K} & \textbf{Math500} & \textbf{Countdown} & \textbf{Sudoku} \\
\midrule
Pass@1 (Student) & 81.3 & 36.8 & 18.2 & 7.2 \\
Pass@8 & 95.5 & 53.6 & 59.2 & 27.7\\
Self-Teacher \\
\quad $\rho_\text{teacher} = 0.10 $ & 85.6 & 40.8 & 29.6 & 11.0 \\
\quad $\rho_\text{teacher} = 0.25 $ & 93.4 & 46.2 & 45.4 & 13.5 \\
\quad $\rho_\text{teacher} = 0.50 $ & 94.8 & 47.4 & 48.2 & 15.4 \\
\bottomrule
\label{tab1}
\end{tabular}
\vspace{-15pt}
\end{wraptable}

\textbf{Training Details. } Following \citep{zhao2026self}, we fix the teacher policy to the initial policy to stabilize training. We use full-vocabulary logit distillation with LoRA \citep{hu2022lora}. The default distribution divergence measure is reverse KL. The generation length and retaining ratio $\rho_\text{teacher}$ are set to $256$ and $0.25$, respectively. Additional training details are provided in \Cref{app4.1}.

\textbf{Evaluation Details. } We evaluated every $25$ steps before step $501$ and report the best results. For mathematical reasoning tasks, we evaluate model performance using generation lengths of $512$ and $256$. For planning tasks, we evaluate at $128$ and $256$. This distinction is made because longer generation lengths improve performance in mathematical reasoning tasks but degrade it in planning tasks (see \Cref{tab2}). We utilize the block diffusion strategy \citep{arriola2025block} with a block length of $32$. Denoising steps are configured as half of the generation length. 

\textbf{Toy Verification. } A critical question that must be answered before the full experiment is whether the self-teacher is strong enough to guide distillation. To verify this, we randomly sampled $500$ questions from each task's training set, obtained generations from the base model, constructed self-teacher inputs (using Pass@$8$) as described in \Cref{sec3.3} under different retaining ratios $\rho_\text{teacher}$, and finally re-generated responses conditioned on these self-teacher inputs. As shown in \Cref{tab1}, even with a moderate $\rho_\text{teacher} = 0.10$, the self-teacher significantly outperforms the student. At higher $\rho_\text{teacher}$, the self-teacher performance nearly matches its origin (Pass@8). This toy experiment successfully validates that our self-teacher can resume correct answers and guide high-quality distillation. Additional details and examples of this toy experiment are provided in \Cref{app4.2}.  

\subsection{Main Results} \label{sec4.2}

\Cref{tab2} presents a comprehensive performance comparison between SFT, RLVR, and our approach. d-OPSD consistently outperforms or matches SFT and RLVR baselines, achieving state-of-the-art performance in most settings and showcasing significant improvements over the base models. \Cref{tab3} and \Cref{fig1} detail the sample efficiency comparison between the RLVR baseline and our approach. d-OPSD demonstrates vastly superior sample efficiency, converging in only around $10\%$ of the optimization steps (number of gradient updates) required by RLVR. Note that the pass@$k$ sampling strategy we use in \Cref{sec3.3} shares the same computation overhead as RLVR (group $k$ rollouts) for each optimization step. Consistent with \citep{lu2025onpolicydistillation, zhao2026self}, we attribute OPSD’s superior sample efficiency to the dense supervision provided by the teacher distribution. These results underscore our approach's promising reasoning performance and sample efficiency.

\subsection{Comparison with AR-style OPSD: Unlocking New Knowledge} \label{sec4.3}

\begin{wraptable}{r}{0.45\textwidth}
\vspace{-20pt}
\setlength{\tabcolsep}{4pt}
\centering
\caption{Reasoning performance Comparison between AR-style OPSD and our approach. Generation length is $256$. Our teacher construction outperforms the AR-style baseline.}
\begin{tabular}{lcc}
\toprule
\textbf{Method / Tasks} & \textbf{GSM8K} & \textbf{Math500} \\
\midrule
LLaDA-8B-Instruct & 76.0 & 31.8 \\
AR-style OPSD & 78.4 & 33.4 \\
\textbf{d-OPSD (Ours)} & \cellcolor{lightblue} 81.0 & \cellcolor{lightblue}  37.2 \\
\bottomrule
\label{tab4}
\end{tabular}
\vspace{-15pt}
\end{wraptable}

A pivotal design choice in our approach is the specific self-teacher construction tailored for dLLMs (\Cref{sec3.1}). It is imperative to evaluate how this formulation compares to the AR-style construction shown in \Cref{fig2}. To this end, we conducted an additional AR-style baseline strictly following \citep{zhao2026self}, which appends the reference solution to the prompt as a prefix conditioning to provide privileged information to the teacher, while keeping our step-level divergence supervision (\Cref{sec3.2}) constant. \Cref{tab4} \footnote{Following \citep{zhao2026self}, the reference solution is the reasoning trajectory obtained directly from the dataset. Therefore, we did not conduct experiments on Countdown and Sudoku, as they consist of only questions and pure ground truths without any reasoning traces.} reports the performance comparison results. Our approach consistently outperforms the AR-style counterpart, highlighting the critical importance of our specific self-teacher construction.

\begin{wrapfigure}{r}{0.4\textwidth}
\vspace{-20pt}
\centering
\includegraphics[width=0.4\textwidth]{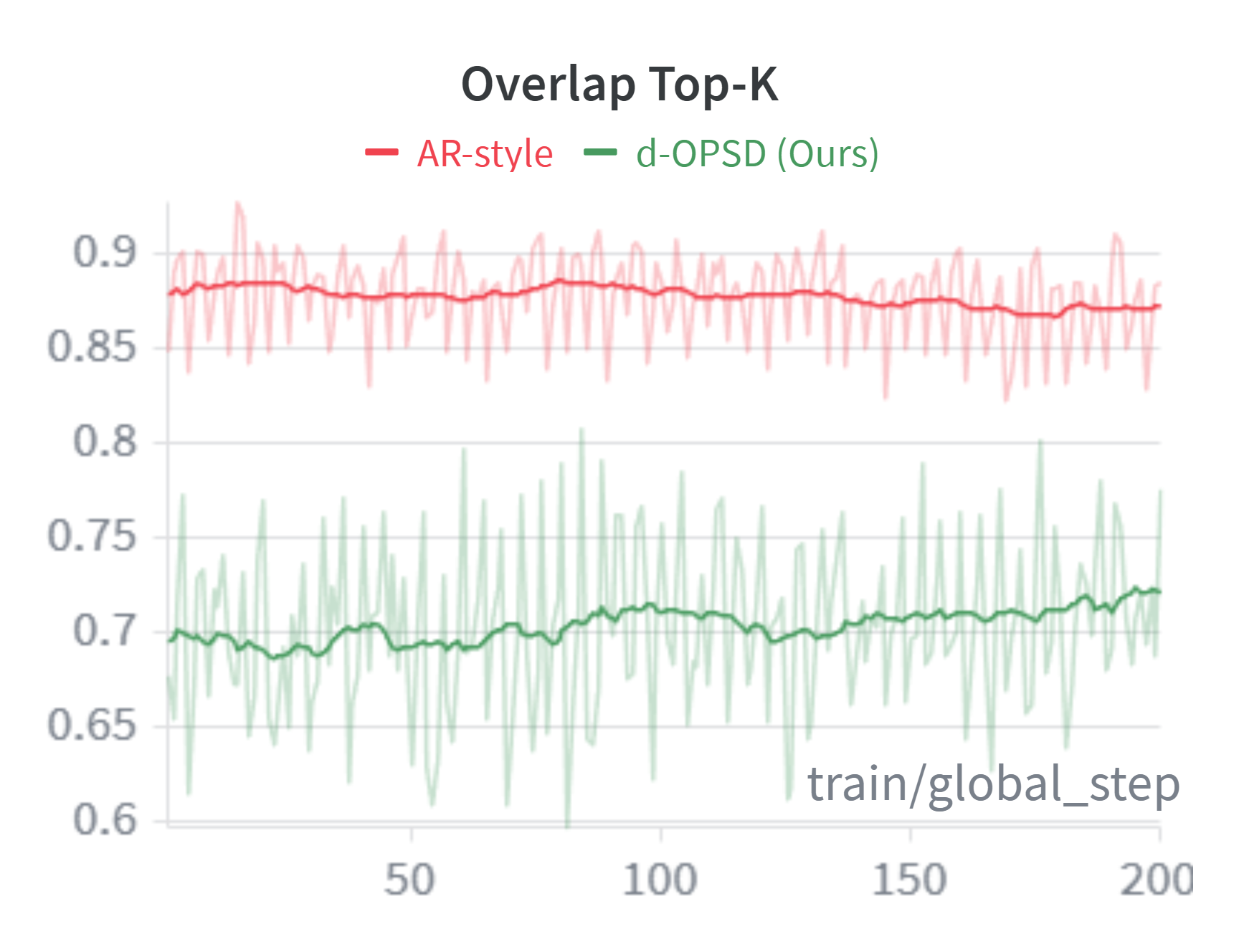}
\vspace{-20pt}
\caption{The Overlap Top-K comparison between d-OPSD and the AR-style counterpart.}
\label{fig3}
\vspace{-15pt}
\end{wrapfigure}

We further investigate the mechanism behind this performance gap. We define the metric of \textit{Overlap Top-$K_t$}. At each denoising step $t$, it measures the proportion of tokens that appear simultaneously in both the student’s and teacher’s Top-$K$ vocabulary distributions over the top-$k$ subset $\mathcal{K}_t$ masked positions. Note that Top-$K$ and top-$k$ have different meanings. Top-$K$ refers to comparing the distribution over the vocabulary at a specific token position, while top-$k$ refers to the most confident tokens in the currently masked positions (\Cref{sec3.2}). Formally, Overlap Top-$K_t$ can be expressed as:
\begin{equation}
    \mathcal{M}_{\text{overlap}, K, t} =  \frac{1}{|\mathcal{K}_t|}  \sum_{i \in \mathcal{K}_t} \left[ \frac{|\mathcal{P}_{\text{student}, t}^{i,\text{Top-}K} \cap \mathcal{P}_{\text{teacher}, t}^{i, \text{Top-}K}|}{K}   \right],
\end{equation}

\begin{wraptable}{r}{0.35\textwidth}
\vspace{-20pt}
\setlength{\tabcolsep}{4pt}
\centering
\caption{Reasoning performance comparison of divergence objectives.}
\begin{tabular}{lc}
\toprule
\textbf{Method / Tasks} & \textbf{GSM8K} \\
\midrule
LLaDA-8B-Instruct & 76.0 \\
Forward KL & 77.9 \\
Reverse KL (default) & \cellcolor{lightblue} 81.0 \\
\bottomrule
\label{tab5}
\end{tabular}
\vspace{-20pt}
\end{wraptable}

where $\mathcal{P}_{t}^{i,\text{Top-}K}$ is the Top-$K$ distribution over the vocabulary at token position $i$, derived from \Cref{eq10}. As shown in Figure 3, the Overlap Top-$K_t$ for AR-style OPSD is extremely high, nearly to $1$, indicating that appending a reference solution fails to bring new knowledge or thinking patterns to the teacher for the student to learn. Conversely, the Overlap Top-$K_t$ for d-OPSD lies in a suitable range, providing more new knowledge that can be transferred from teacher to student. $K$ is set to $K=20$ in practice.

\subsection{Ablation Studies} \label{sec4.4}

Additional ablation studies are provided in \cref{app5}.

\begin{wraptable}{r}{0.38\textwidth}
\vspace{-25pt}
\setlength{\tabcolsep}{4pt}
\centering
\caption{Reasoning performance comparison of retaining ratios.}
\begin{tabular}{lc}
\toprule
\textbf{Method / Tasks} & \textbf{GSM8K} \\
\midrule
LLaDA-8B-Instruct & 76.0  \\
diffu-GRPO & 79.8 \\
d-OPSD \\
\quad $\rho_\text{teacher}=0.10$ & 80.5 \\
\quad $\rho_\text{teacher}=0.25$ (default) & \cellcolor{lightblue} 81.0 \\
\quad $\rho_\text{teacher}=0.50$ & 79.8 \\
\bottomrule
\label{tab6}
\end{tabular}
\vspace{-25pt}
\end{wraptable}

\textbf{Divergence Objective. } We compare reverse KL (default) and forward KL in \Cref{tab5}. Reverse KL clearly outperforms forward KL. We attribute this to the model-seeking behavior of reverse KL \citep{agarwal2024policy}, which is more robust compared to the model-covering behavior of forward KL.

\textbf{Retaining Ratio. } We observe that different retaining ratios $\rho_\text{teacher}$ have moderate influences on overall performance. As shown in \Cref{tab6}, all configurations improve over the base model and surpass the RLVR baseline. Interestingly, $\rho_\text{teacher}=0.10$ yields better results than $\rho_\text{teacher}=0.50$, despite it is a weaker teacher as shown in \Cref{tab1}. This suggests that while a accurate teacher is beneficial, the distillation effectiveness is not only decided by the teacher performance.

\begin{wraptable}{r}{0.35\textwidth}
\vspace{-15pt}
\setlength{\tabcolsep}{4pt}
\centering
\caption{Reasoning performance comparison of $\mathcal{K}_t$ selections.}
\begin{tabular}{lc}
\toprule
\textbf{Method / Tasks} & \textbf{GSM8K} \\
\midrule
LLaDA-8B-Instruct & 76.0 \\
From Student & 78.6 \\
From Teacher (default) & \cellcolor{lightblue} 81.0 \\
\bottomrule
\label{tab7}
\end{tabular}
\vspace{-20pt}
\end{wraptable}

\textbf{top-$k$ subset $\mathcal{K}_t$ Selection. } \label{sec4.4.3} As noted in \cref{sec3.2}, $\mathcal{K}_t$ can be selected using either the student distribution or teacher distribution. \Cref{tab7} compares these two choice. Deriving $\mathcal{K}_t$ from the teacher distribution yields higher performance, as it forces the student to align with the most confident distributions by the teacher policy, providing a stronger learning signal.

\textbf{Pass@$k$. } \label{sec4.4.4} As noted in \Cref{sec3.3}, we employ a sampling strategy akin to pass@$k$, keeping sampling trajectories until a correct answer occurs within $k$ iterations. 

\begin{wraptable}{r}{0.45\textwidth}
\vspace{-20pt}
\setlength{\tabcolsep}{4pt}
\centering
\caption{Reasoning performance comparison of sampling strategies.}
\begin{tabular}{lcc}
\toprule
\textbf{Method / Tasks} & \textbf{GSM8K} & \textbf{Countdown} \\
\midrule
LLaDA-8B-Instruct & 76.0 & 20.3 \\
diffu-GRPO & 79.8 & 33.2 \\
d-OPSD \\
\quad $k=1$ & 80.4 & 34.0 \\
\quad $k=8$ (default) & \cellcolor{lightblue} 81.0 & \cellcolor{lightblue} 37.9 \\
\bottomrule
\label{tab8}
\end{tabular}
\vspace{-25pt}
\end{wraptable}

\Cref{tab8} evaluates the impact of varying $k$. Although $k=1$ slightly degrades reasoning performance compared to $k=8$, it still surpasses the RLVR baseline with a even greater sample efficiency than $k=8$.

\begin{wraptable}{r}{0.35\textwidth}
\vspace{-20pt}
\setlength{\tabcolsep}{4pt}
\centering
\caption{Reasoning performance comparison of clipping.}
\begin{tabular}{lc}
\toprule
\textbf{Method / Tasks} & \textbf{GSM8K} \\
\midrule
LLaDA-8B-Instruct & 76.0 \\
No-clip & 77.0 \\
Clip (default) & \cellcolor{lightblue} 81.0 \\
\bottomrule
\label{app_tab3}
\end{tabular}
\vspace{-20pt}
\end{wraptable}

\textbf{Per-token Pointwise Clipping. } As noted in \Cref{app3.1}, we adopt a pointwise clipping strategy following \citep{zhao2026self}. \Cref{app_tab3} shows that pointwise clipping substantially improves the performance of d-OPSD. More importantly, we observe that clipping stabilizes training in most settings, which explains the performance gap. In contrast, the none-clipping variant starts to collapse around step $150$, with performance finally dropping to $69.37$ by step $500$. The clipping threshold is set to $0.05$ in practice.

\subsection{Failure Modes} \label{sec4.5}

We wish to transparently share a failure mode observed with our current approach. Although it is highly effective in both reasoning performance and sample efficiency, we find that similar to RLVR, OPSD in some settings is prone to policy collapse after achieving peak performance. 

As shown in \Cref{fig4}, training sometimes degrade catastrophically. We noticed that the same phenomena is commonly observed in RLVR \citep{deng2025grpo, bai2025m}. We hypothesize that this collapse may stem from the model-seeking behavior \citep{agarwal2024policy} becoming overly narrow, prevent from further learning.

\section{Related Works}

Additional related works are provided in \Cref{app1.2}.

\textbf{On-policy Distillation. } Knowledge distillation \citep{hinton2015distilling} transfers knowledge from a large teacher model to a smaller student model by training on the teacher's soft output distributions. \citet{kim2016sequence, jiao2020tinybert, wang2020minilm} leveraged it to sequence-level distillation, establishing the dominant off-policy distillation approaches. \citep{gu2024minillm, agarwal2024policy, lu2025onpolicydistillation, yang2026learning} extended it to OPD, addressing the exposure bias \citep{bengio2015scheduled} mismatch by shifting the training distribution to the student's own generations.

\section{Conclusion} \label{sec6}

This work presents d-OPSD, the first on-policy self-distillation approach for dLLMs. It is specifically tailored to align with on-policy and dLLMs nature. We propose a novel self-teacher construction that utilizes the model's own self-generated answers as suffix conditioning for privileged information, effectively guiding the student to learn from its on-policy ``self-future experience''. Furthermore, we shift the dense divergence supervision from the token-level to step-level, perfectly matching the iterative mechanics of dLLMs. Future work will explore advanced techniques to further stabilize and enhance the OPSD post-training of dLLMs.

\bibliographystyle{unsrtnat}
\bibliography{neurips_2026}

% \medskip

% {
% \small

% [1] Alexander, J.A.\ \& Mozer, M.C.\ (1995) Template-based algorithms for
% connectionist rule extraction. In G.\ Tesauro, D.S.\ Touretzky and T.K.\ Leen
% (eds.), {\it Advances in Neural Information Processing Systems 7},
% pp.\ 609--616. Cambridge, MA: MIT Press.

% [2] Bower, J.M.\ \& Beeman, D.\ (1995) {\it The Book of GENESIS: Exploring
%   Realistic Neural Models with the GEneral NEural SImulation System.}  New York:
% TELOS/Springer--Verlag.

% [3] Hasselmo, M.E., Schnell, E.\ \& Barkai, E.\ (1995) Dynamics of learning and
% recall at excitatory recurrent synapses and cholinergic modulation in rat
% hippocampal region CA3. {\it Journal of Neuroscience} {\bf 15}(7):5249-5262.
% }

%%%%%%%%%%%%%%%%%%%%%%%%%%%%%%%%%%%%%%%%%%%%%%%%%%%%%%%%%%%%
\newpage
\appendix

\section{Additional Preliminaries and Related Works}

\subsection{Additional Preliminaries} \label{app1.1}

\textbf{Block-diffusion. } In practice, the block-diffusion inference strategy \citep{han2023ssd, arriola2025block, fathi2025unifying} is commonly used in current dLLMs. This hybrid approach partitions a response $y$ into $B$ contiguous, non-overlapping blocks $\{\text{block}_1, \text{block}_2, \cdots, \text{block}_B\}$, with each block containing $L' = \frac{L}{B}$ tokens. The inference is purely AR at the block level while being purely diffusion-style within each block, where the next block starts to decode only when the last block gets fully decoded.

\subsection{Additional Related Works} \label{app1.2}

\textbf{Reinforcement Learning for dLLMs.} Reinforcement learning (RL) \citep{guo2025deepseek,shao2024deepseekmath,zhang2026consolidating} has emerged as a critical post-training technique for enhancing the reasoning capabilities of dLLMs. Most existing works \citep{zhao2025d1, huang2025reinforcing, tang2025wd1, gong2025diffucoder, rojas2025improving, ou2025principled} directly apply GRPO to dLLMs, using either one-step estimation or the ELBO to estimate the log-probability in GRPO. However, most of them suffer from the fundamental challenges of RLVR: the heavy computation overhead and the bottleneck of spare rewards.

\textbf{Self-distillation for dLLMs. } Recently there are some self-distillation relevant works for dLLMs, however our method is fundamentally distinct from them. GSSD \citep{tang2026gdsd} and DiPOD \citep{jiang2026dipod} remain as RLVR methods, as they still fall under the policy-gradient RLVR framework using sparse verifiable rewards to improve the policy. In this framework, self-distillation acts merely as an auxiliary component. Specifically, GDSD uses off-policy self-distillation to replace the likelihood term in RLVR, and DiPOD uses self-distillation as an additional regularization in RLVR to tight the bounded policy. In contrast, our method is a pure OPSD method which provides dense KL-divergence supervision to push the student closer to the privileged teacher, without relying on the sparse rewards or policy-gradient paradigms. OPDLM \citep{su2026data} focuses on converting the model architecture from an original AR model into a dLLM. Conversely, our method is designed to enhance the reasoning capabilities of original dLLMs. Our method differs from abovementioned works by proposing the first pure OPSD framework for enhancing dLLMs' reasoning capabilities.

\section{Self-teacher Construction Illustrations} \label{app2}

Here we provide an example of how our self-teacher construction (\Cref{sec3.1}) works, with a question sampled from GSM8K training set. For brevity, we omit some $\mask$ and ``end-of-text'' tokens.

The question is:

\begin{figure}[htbp]
\vspace{-5pt}
\centering{\includegraphics[scale=0.4]{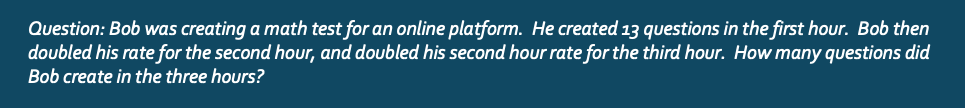}}
\vspace{-2pt}
\caption{A question from GSM8K training set.}
\vspace{-5pt}
\end{figure}

\begin{figure}[htbp]
\vspace{-8pt}
\centering{\includegraphics[scale=0.31]{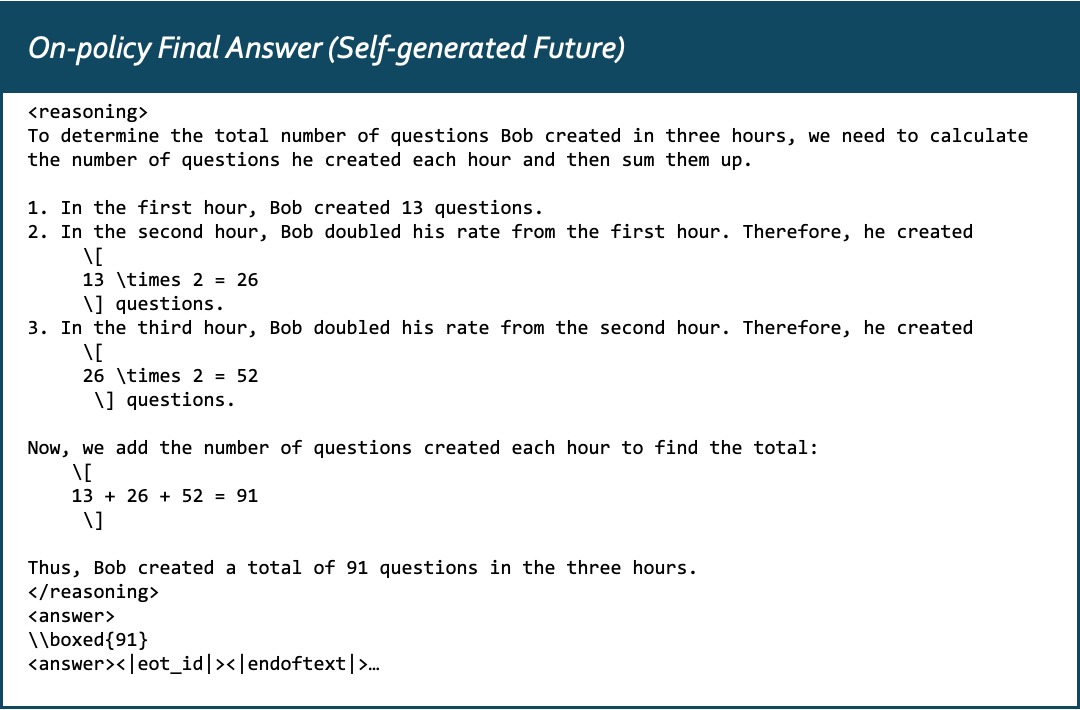}}
\vspace{-2pt}
\caption{The self-generated future answer.}
\label{fig5}
\end{figure}

\begin{figure}[htbp]
\vspace{-5pt}
\centering{\includegraphics[scale=0.35]{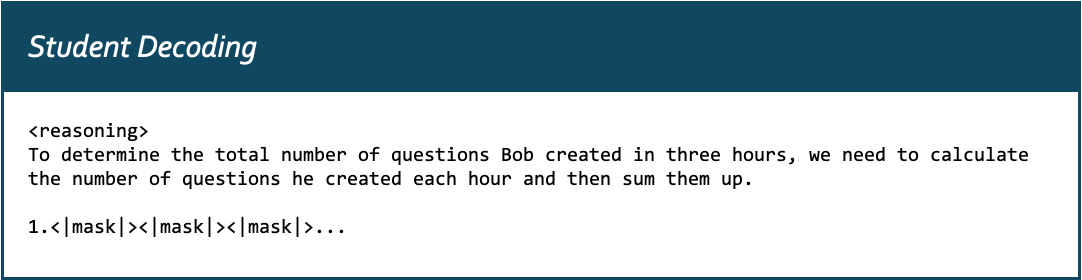}}
\vspace{-2pt}
\caption{Current student decoding status.}
\vspace{-5pt}
\label{fig6}
\end{figure}

\begin{figure}[htbp]
\vspace{-5pt}
\centering{\includegraphics[scale=0.35]{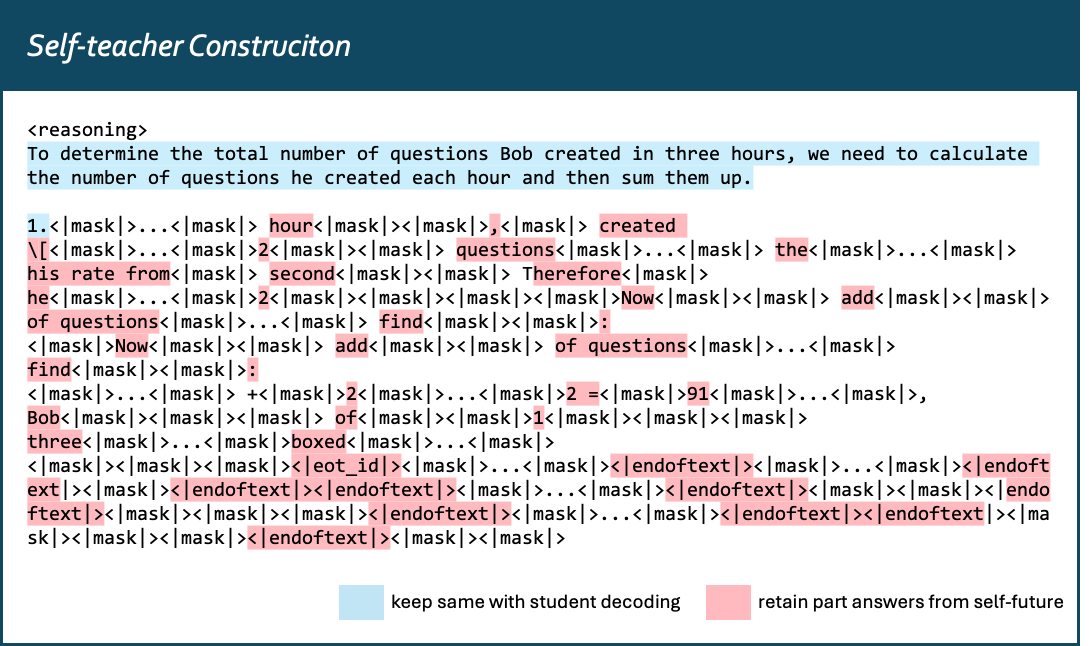}}
\vspace{-2pt}
\caption{Self-teacher construction at $t=20$.}
\vspace{-5pt}
\label{fig7}
\end{figure}

First, we sample an on-policy trajectory \footnote{Using pass@$k$, it keeps sampling until a correct final answer appears or it reaches the iteration threshold.} from the student model and obtain the final clean answer as the self-generated future, as shown in \Cref{fig5}. Specifically, at denoising step $t=20$, we have the student decoding status as shown in \Cref{fig6}. We then construct the self-teacher at step $t=20$ as shown in \Cref{fig7}.

\begin{figure}[htbp]
\vspace{-5pt}
\centering{\includegraphics[scale=0.35]{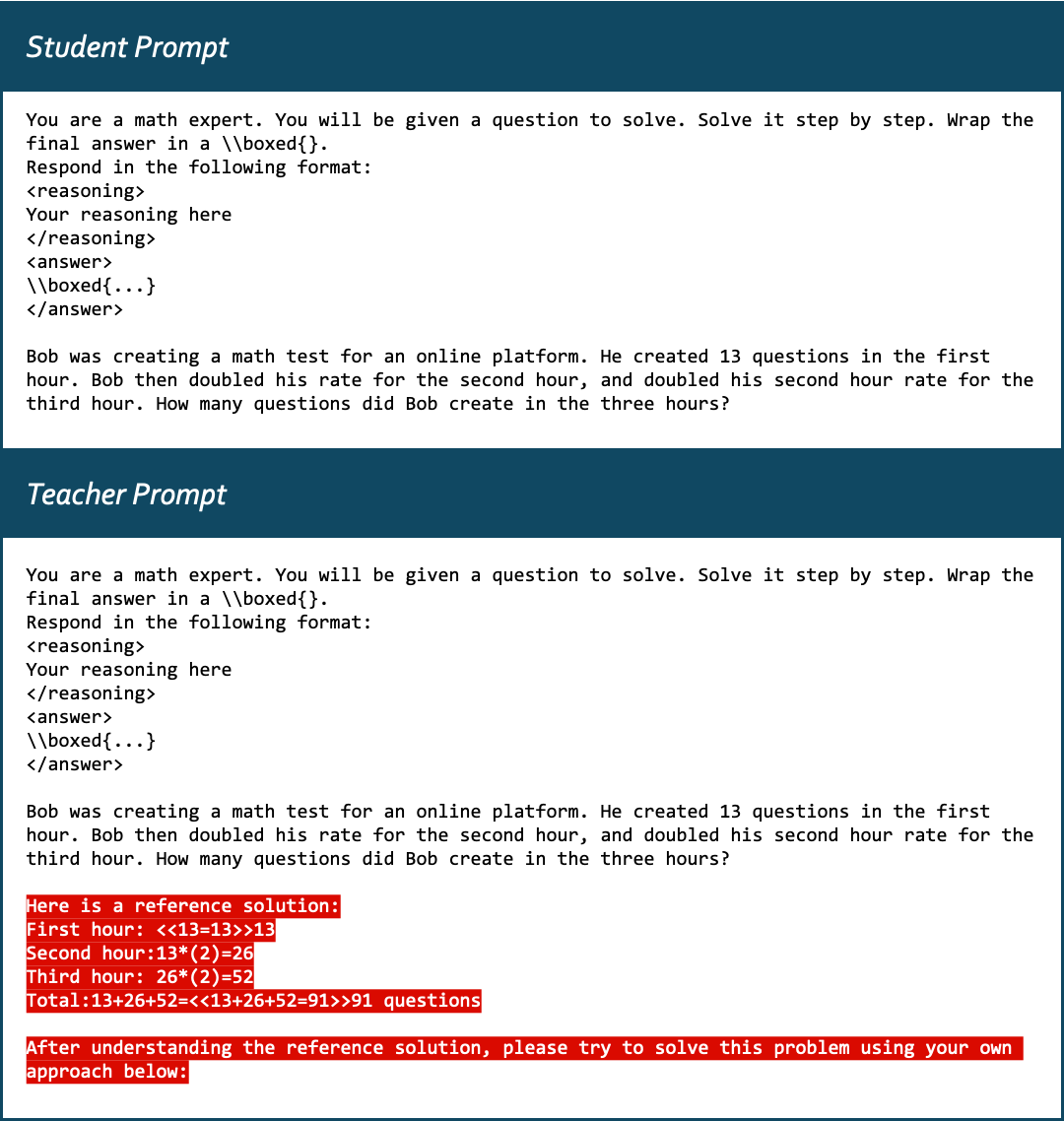}}
\vspace{-2pt}
\caption{AR-style teacher construction.}
\vspace{-5pt}
\label{fig_app5}
\end{figure}

For comparison, we also illustrate the AR-style construction, which appends a reference solution to the prompt, as shown in \Cref{fig_app5}.

\section{Additional Implementation Details} \label{app3}

\subsection{Per-Token pointwise clipping} \label{app3.1}

Following \citep{zhao2026self}, we apply pointwise clipping to the vocabulary level divergence contributions. The reason is that  token-level divergence is highly skewed across vocabulary entries, and our ablation study in \Cref{sec4.4} empirically validates that pointwise clipping stabilizes training and leads to better performance.

\subsection{Inputs Concatenation}

DLLMs generate responses by an iterative denoising process, where each iteration requires full-attention over all token positions. Consequently, computing the loss objective in \Cref{eq12} can easily lead to out of memory, as the full attention gradient maps over all token positions across every steps need to be stored, until a trajectory is fully decoded. To address this issue, we leverage a engineering technique, concatenating all inputs across every steps of a trajectory into an entire batch. Specifically, assume that the student decoding status is a tensor of shape (bsz, seq-length). Instead of feeding it into the model to compute the corresponding term in \Cref{eq12}, we concatenate all status tensors across all steps of this trajectory to form a ``batch'' tensor of shape (bsz$\times$steps, seq-length). Since all inputs share the same model, the gradient remains constant for each input and no longer needs to be stored as previously.

\subsection{Compute only on Correct Generations}

By default, we compute the loss objective \Cref{eq12} only on correct generations \footnote{For Sudoku task, there are no ``right'' or ``wrong'' answers because it gives a score in $[0, 1]$. Therefore, we set an threshold to decide if the generation should be include in loss computation. In practice, the threshold is set to $0.25$.}. Although computing on all generations also improves the model's reasoning performance, our default setting achieves superior results. Detailed experimental results are provided in \Cref{app5.1}.

\section{Additional Experiment Details}

\subsection{Training Details} \label{app4.1}

We used the TRL library \citep{vonwerra2020trl} to implement d-OPSD. We employed Low-Rank Adaptation (LoRA) with a rank of $r = 128$ and scaling factor $\alpha = 64$. Training was conducted on $4$ NVIDIA GPUs, with a learning rate of $5 \times 10^{-6}$, accumulation steps of $1$, the AdamW optimizer \citep{loshchilov2017decoupled}, and Flash Attention 2 \citep{dao2023flashattention}. The RLVR baseline diffu-GRPO \citep{zhao2025d1} in \Cref{fig1} was reproduced on $8$ NVIDIA GPUs, following its default settings.

\subsection{Toy Experiment Details and Examples} \label{app4.2}

The generation length is $256$ for all tasks. After applying the self-teacher construction, the number of remaining $\mask$ tokens becomes smaller than the generation length $256$. We keep constant that unmasking $2$ tokens in each step with a block length of $32$.

One important point to note is that, to prevent the risk of leaking the final answer (e.g., the final answer between <answer><answer> is retained in the self-teacher construction), everytime we move to a new block, we clear all unmasked tokens in this block, leaving the new block entirely filled with only $\mask$ tokens. 

We provide an example from GSM8K training set. The question is:

\begin{figure}[htbp]
\vspace{-5pt}
\centering{\includegraphics[scale=0.4]{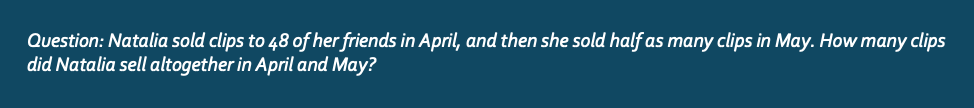}}
\vspace{-2pt}
\caption{A question from GSM8K training set.}
\vspace{-5pt}
\end{figure}

First, we sample a generation \footnote{Using pass@$k$, it keeps sampling until a correct final answer appears or it reaches the iteration threshold.} from the student model and obtain the final clean answer:

\begin{figure}[htbp]
\vspace{-8pt}
\centering{\includegraphics[scale=0.36]{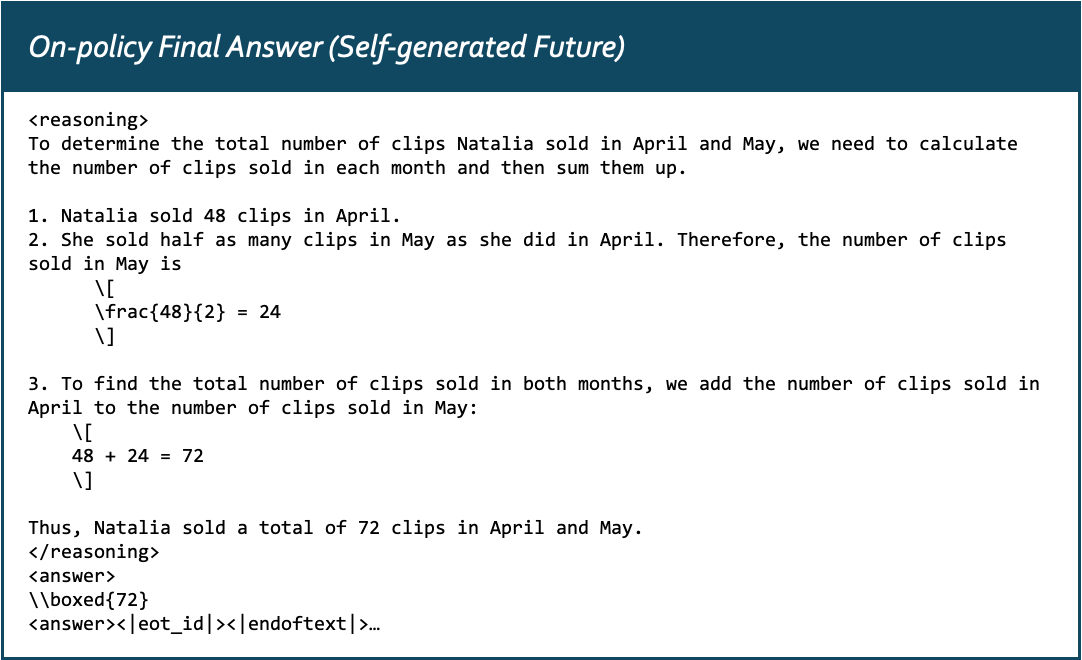}}
\vspace{-2pt}
\caption{The self-generated answer.}
\end{figure}

\begin{figure}[htbp]
\vspace{-8pt}
\centering{\includegraphics[scale=0.36]{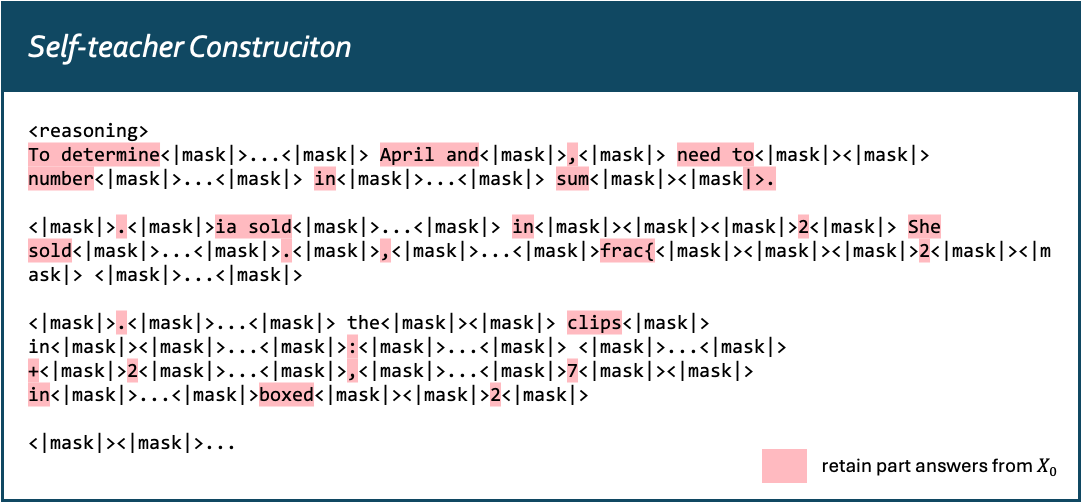}}
\vspace{-2pt}
\caption{Self-teacher in the toy experiment.}
\vspace{-14pt}
\label{fig_app8}
\end{figure}

We then construct self-teacher by partially revealing the final generation, as shown in \Cref{fig_app8}.

\begin{wraptable}{r}{0.39\textwidth}
\vspace{-65pt}
\setlength{\tabcolsep}{4pt}
\centering
\caption{Reasoning performance comparison of teacher fixing.}
\begin{tabular}{lc}
\toprule
\textbf{Method / Tasks} & \textbf{GSM8K} \\
\midrule
LLaDA-8B-Instruct & 76.0 \\
diffu-GRPO & 79.8 \\
d-OPSD \\
\quad Unfix teacher & 79.7 \\
\quad Fix teacher (default) & \cellcolor{lightblue} 81.0 \\
\bottomrule
\label{app_tab1}
\end{tabular}
\vspace{-30pt}
\end{wraptable}

\section{Additional Experiment Results} \label{app5}

\subsection{Additional Ablation Studies} \label{app5.1}

\textbf{Fixing the Teacher. } We find that fixing the teacher model leads to greater performance gains, as shown in \Cref{app_tab1}. Notably, even when the teacher is not fixed, d-OPSD's reasoning performance nearly matches the RLVR baselines, further demonstrating its effectiveness.

\newpage

\begin{wraptable}{r}{0.38\textwidth}
\vspace{-7pt}
\setlength{\tabcolsep}{4pt}
\centering
\caption{Reasoning performance comparison.}
\begin{tabular}{lc}
\toprule
\textbf{Method / Tasks} & \textbf{GSM8K} \\
\midrule
LLaDA-8B-Instruct & 76.0 \\
diffu-GRPO & 79.8 \\
d-OPSD \\
\quad All & 80.3 \\
\quad Correct only (default) & \cellcolor{lightblue} 81.0 \\
\bottomrule
\label{app_tab2}
\end{tabular}
\vspace{-30pt}
\end{wraptable}

\textbf{Compute only on Correct Generations. } \label{app5.2} As shown in \Cref{app_tab2}, computing the loss on all trajectories leads to a slight performance degradation. Nevertheless, it still outperforms the RLVR baseline.

\textbf{Self-SFT Baseline. } When computing only on correct generations, our method is related to verifier-guided rejection-sampling self-training. To make a fair comparison, we also provide a step-level rejection-sampling self-SFT baseline strictly using the same settings. The results in \Cref{tab12} demonstrate that our method significantly outperforms the baseline across all tasks.

More deeply, we would like to clarify that, the performance gap above lies in the fundamental difference between our method and the rejection-sampling self-SFT. By using self-generated answers to construct the self-teacher, our method provides a dense KL-divergence supervision instead of hard labels. This is aligned with our ablations in \Cref{app_tab2}, where we trained our method on all trajectories without any rejection sampling and it still outperforms the RLVR and self-SFT baselines, confirming that our method still works without rejection sampling. This result further demonstrates that the dense distillation mechanism is the primary driver of our method's success.

\begin{table}[!t]
\centering
\caption{Reasoning performance comparison with the Self-SFT baseline.}
\begin{threeparttable}
\begin{adjustbox}{max width=\textwidth}
\begin{tabular}{lcccccccc}
\toprule
\multirow{2}{*}{Model / Seq Len} & \multicolumn{2}{c}{GSM8K} & \multicolumn{2}{c}{MATH500} & \multicolumn{2}{c}{Countdown} & \multicolumn{2}{c}{Sudoku} \\
\cmidrule(lr){2-3} \cmidrule(lr){4-5} \cmidrule(lr){6-7} \cmidrule(lr){8-9}
 & 256 & 512 & 256 & 512 & 128 & 256 & 128 & 256 \\
\midrule

LLaDA-8B-Instruct & 76.0 & 79.5 & 31.8 & 36.2 & 20.3 & 19.1 & 11.5 & 6.9 \\
\midrule
Self-SFT & 72.2 & 80.7 & 31.8 & 37.2 & 16.0 & 29.8 & 9.7 & 13.7 \\
\textbf{d-OPSD (Ours)} & \cellcolor{lightblue} 81.0 & \cellcolor{lightblue} 82.2 & \cellcolor{lightblue} 37.2 & 37.8 & \cellcolor{lightblue} 37.9 & \cellcolor{lightblue} 32.3 & \cellcolor{lightblue} 23.9 & \cellcolor{lightblue} 20.6 \\

\bottomrule
\label{tab12}
\end{tabular}
\end{adjustbox}
\end{threeparttable}
\vspace{-5pt}
\end{table}

\begin{table}[!t]
\centering
\caption{Reasoning performance comparison in full-finetuning.}
\begin{threeparttable}
\begin{adjustbox}{max width=\textwidth}
\begin{tabular}{lcccccccc}
\toprule
\multirow{2}{*}{Model / Seq Len} & \multicolumn{2}{c}{GSM8K} & \multicolumn{2}{c}{MATH500} & \multicolumn{2}{c}{Countdown} & \multicolumn{2}{c}{Sudoku} \\
\cmidrule(lr){2-3} \cmidrule(lr){4-5} \cmidrule(lr){6-7} \cmidrule(lr){8-9}
 & 256 & 512 & 256 & 512 & 128 & 256 & 128 & 256 \\
\midrule

LLaDA-8B-Instruct & 76.0 & 79.5 & 31.8 & 36.2 & 20.3 & 19.1 & 11.5 & 6.9 \\
\midrule
Full-finetuning \\
\quad diffu-GRPO & 78.5 & 77.6 & 34.8 & 36.0 & 28.5 & \cellcolor{lightblue} 26.3 & 22.0 & 19.2 \\
\quad \textbf{d-OPSD (Ours)} & \cellcolor{lightblue} 79.6 & \cellcolor{lightblue} 80.7 & \cellcolor{lightblue} 36.0 & 36.8 & \cellcolor{lightblue} 36.7 & 25.0 & \cellcolor{lightblue} 23.6 & \cellcolor{lightblue} 22.0 \\

\bottomrule
\label{tab13}
\end{tabular}
\end{adjustbox}
\end{threeparttable}
\vspace{-20pt}
\end{table}

\textbf{Full-finetuning Baseline. } We also provide a full finetuning baseline. Results in \Cref{tab13} demonstrate that under the full fine-tuning setting, our approach consistently outperforms the RLVR method across most of the tasks. This further validates the effectiveness and robustness of our approach.

\subsection{Qualitative Examples on GSM8k}

We provide a qualitative example from GSM8k testing set, where the RLVR model gives an incorrect answer while our approach yields the correct one, as shown in \Cref{fig_app9}.

\subsection{Failure Mode}

\begin{figure}[htbp]
\vspace{45pt}
\centering{\includegraphics[scale=0.32]{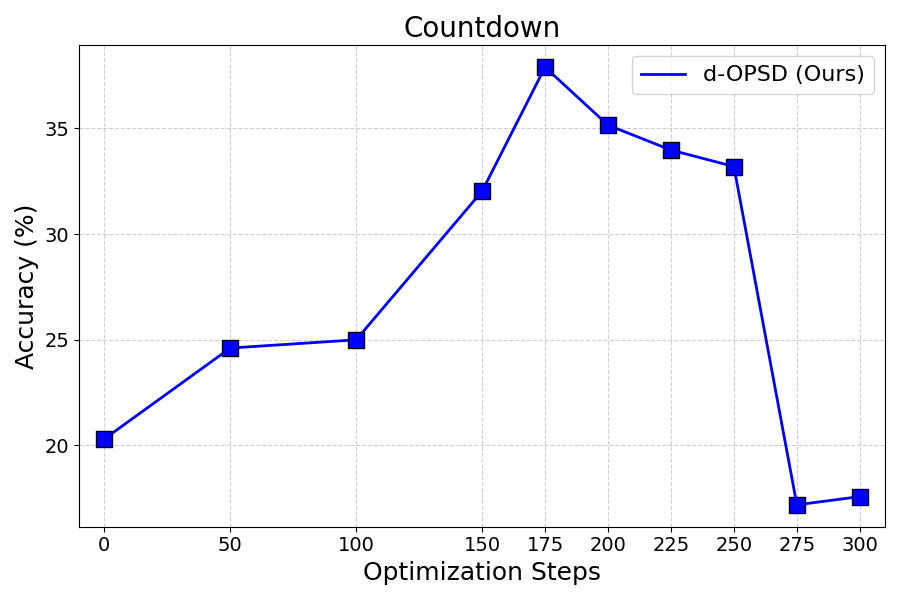}}
\vspace{-2pt}
\caption{Failure Mode of collapse.}
\label{fig4}
\end{figure}

\begin{figure}[htbp]
\vspace{-8pt}
\centering{\includegraphics[scale=0.36]{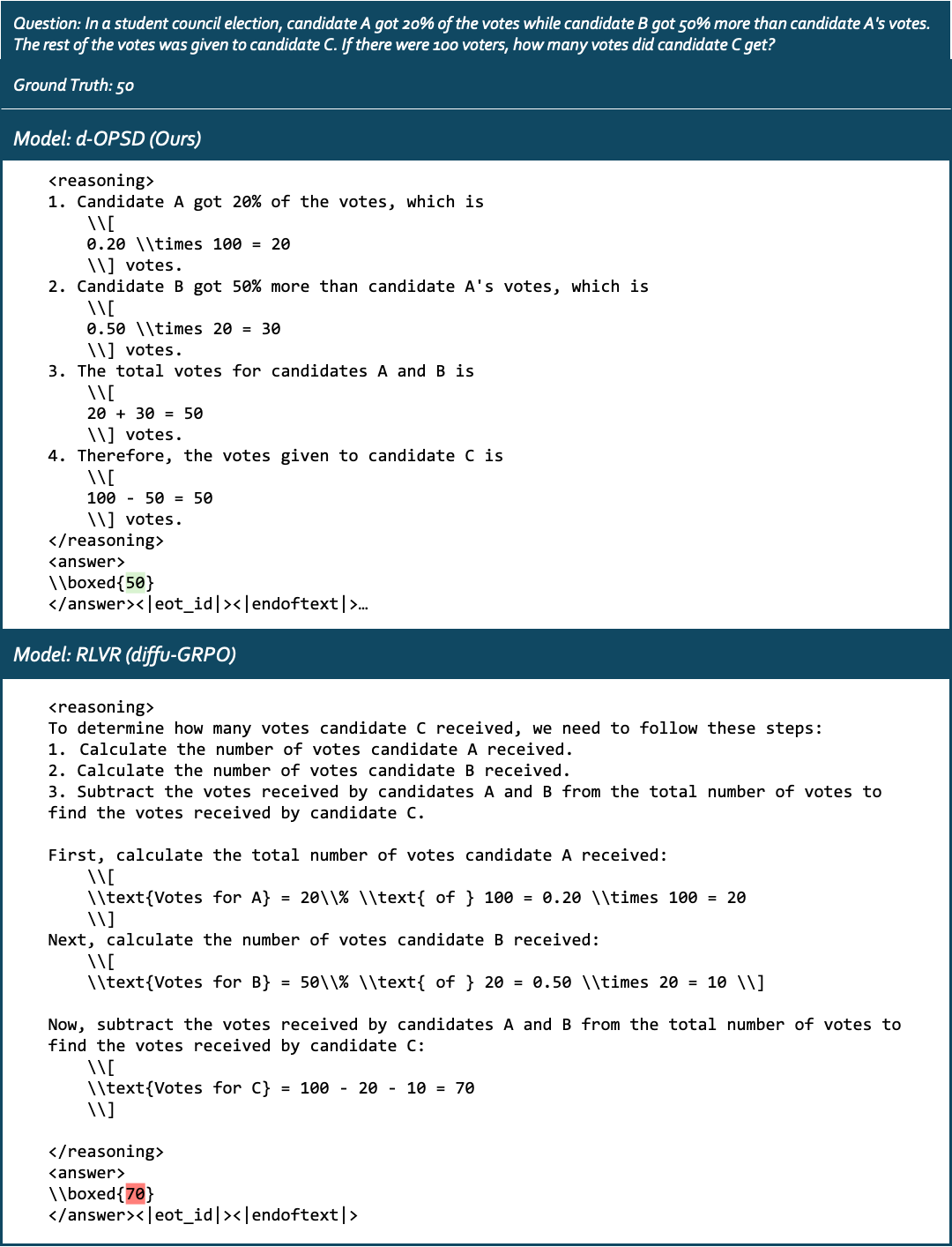}}
\vspace{-2pt}
\caption{Qualitative Examples on GSM8k}
\label{fig_app9}
\end{figure}

\Cref{fig4} presents the failure mode mentioned in \Cref{sec4.5}.

%%%%%%%%%%%%%%%%%%%%%%%%%%%%%%%%%%%%%%%%%%%%%%%%%%%%%%%%%%%%

\newpage
\section*{NeurIPS Paper Checklist}

\begin{enumerate}

\item {\bf Claims}
    \item[] Question: Do the main claims made in the abstract and introduction accurately reflect the paper's contributions and scope?
    \item[] Answer: \answerYes{} % Replace by \answerYes{}, \answerNo{}, or \answerNA{}.
    \item[] Justification: The main claims made in the abstract and introduction accurately reflect the paper’s contributions and scope.

\item {\bf Limitations}
    \item[] Question: Does the paper discuss the limitations of the work performed by the authors?
    \item[] Answer: \answerYes{} % Replace by \answerYes{}, \answerNo{}, or \answerNA{}.
    \item[] Justification: See \Cref{sec4.5}.

\item {\bf Theory assumptions and proofs}
    \item[] Question: For each theoretical result, does the paper provide the full set of assumptions and a complete (and correct) proof?
    \item[] Answer: \answerYes{} % Replace by \answerYes{}, \answerNo{}, or \answerNA{}.
    \item[] Justification: We provide details for our theory.

    \item {\bf Experimental result reproducibility}
    \item[] Question: Does the paper fully disclose all the information needed to reproduce the main experimental results of the paper to the extent that it affects the main claims and/or conclusions of the paper (regardless of whether the code and data are provided or not)?
    \item[] Answer: \answerYes{} % Replace by \answerYes{}, \answerNo{}, or \answerNA{}.
    \item[] Justification: See \Cref{sec4.1} and \Cref{app4.1}, and our code with the github link.

\item {\bf Open access to data and code}
    \item[] Question: Does the paper provide open access to the data and code, with sufficient instructions to faithfully reproduce the main experimental results, as described in supplemental material?
    \item[] Answer: \answerYes{} % Replace by \answerYes{}, \answerNo{}, or \answerNA{}.
    \item[] Justification: We provide our code with the github link, see Abstract.

\item {\bf Experimental setting/details}
    \item[] Question: Does the paper specify all the training and test details (e.g., data splits, hyperparameters, how they were chosen, type of optimizer) necessary to understand the results?
    \item[] Answer: \answerYes{} % Replace by \answerYes{}, \answerNo{}, or \answerNA{}.
    \item[] Justification: See \Cref{sec4.1} and \Cref{app4.1}.

\item {\bf Experiment statistical significance}
    \item[] Question: Does the paper report error bars suitably and correctly defined or other appropriate information about the statistical significance of the experiments?
    \item[] Answer: \answerYes{} % Replace by \answerYes{}, \answerNo{}, or \answerNA{}.
    \item[] Justification: See \Cref{sec4}.

\item {\bf Experiments compute resources}
    \item[] Question: For each experiment, does the paper provide sufficient information on the computer resources (type of compute workers, memory, time of execution) needed to reproduce the experiments?
    \item[] Answer: \answerYes{} % Replace by \answerYes{}, \answerNo{}, or \answerNA{}.
    \item[] Justification: See \Cref{app4.1}.
    
\item {\bf Code of ethics}
    \item[] Question: Does the research conducted in the paper conform, in every respect, with the NeurIPS Code of Ethics \url{https://neurips.cc/public/EthicsGuidelines}?
    \item[] Answer:\answerYes{} % Replace by \answerYes{}, \answerNo{}, or \answerNA{}.
    \item[] Justification: The research conducted in the paper conform, in every respect, with the NeurIPS Code of Ethics.

\item {\bf Broader impacts}
    \item[] Question: Does the paper discuss both potential positive societal impacts and negative societal impacts of the work performed?
    \item[] Answer: \answerYes{} % Replace by \answerYes{}, \answerNo{}, or \answerNA{}.
    \item[] Justification: See \Cref{sec6} and \Cref{sec4.5}.
    
\item {\bf Safeguards}
    \item[] Question: Does the paper describe safeguards that have been put in place for responsible release of data or models that have a high risk for misuse (e.g., pre-trained language models, image generators, or scraped datasets)?
    \item[] Answer: \answerNA{} % Replace by \answerYes{}, \answerNo{}, or \answerNA{}.
    \item[] Justification: See \Cref{sec6}.

\item {\bf Licenses for existing assets}
    \item[] Question: Are the creators or original owners of assets (e.g., code, data, models), used in the paper, properly credited and are the license and terms of use explicitly mentioned and properly respected?
    \item[] Answer: \answerYes{} % Replace by \answerYes{}, \answerNo{}, or \answerNA{}.
    \item[] Justification: All creators or original owners of assets (e.g., code, data, models), used in the paper, has been properly credited; the license and terms of use have been explicitly mentioned and properly respected.

\item {\bf New assets}
    \item[] Question: Are new assets introduced in the paper well documented and is the documentation provided alongside the assets?
    \item[] Answer: \answerYes{} % Replace by \answerYes{}, \answerNo{}, or \answerNA{}.
    \item[] Justification: All codes are released.

\item {\bf Crowdsourcing and research with human subjects}
    \item[] Question: For crowdsourcing experiments and research with human subjects, does the paper include the full text of instructions given to participants and screenshots, if applicable, as well as details about compensation (if any)? 
    \item[] Answer: \answerNA{} % Replace by \answerYes{}, \answerNo{}, or \answerNA{}.
    \item[] Justification: \answerNA{}

\item {\bf Institutional review board (IRB) approvals or equivalent for research with human subjects}
    \item[] Question: Does the paper describe potential risks incurred by study participants, whether such risks were disclosed to the subjects, and whether Institutional Review Board (IRB) approvals (or an equivalent approval/review based on the requirements of your country or institution) were obtained?
    \item[] Answer: \answerNA{} % Replace by \answerYes{}, \answerNo{}, or \answerNA{}.
    \item[] Justification: \answerNA{}

\item {\bf Declaration of LLM usage}
    \item[] Question: Does the paper describe the usage of LLMs if it is an important, original, or non-standard component of the core methods in this research? Note that if the LLM is used only for writing, editing, or formatting purposes and does \emph{not} impact the core methodology, scientific rigor, or originality of the research, declaration is not required.
    %this research? 
    \item[] Answer: \answerYes{} % Replace by \answerYes{}, \answerNo{}, or \answerNA{}.
    \item[] Justification: LLMs are used for part of coding, and all LLM-generated codes are checked and reviewed by the authors.

\end{enumerate}

\end{document}